\documentclass{article}

\usepackage{microtype}
\usepackage{graphicx}
\usepackage{subfigure}
\usepackage{booktabs} 

\usepackage{hyperref}
\usepackage{tabularx}

\newcommand{\nl}{\textcolor{gray}{\textbackslash n}}
\newcommand{\param}[1]{\textcolor{purple}{\{#1\}}}
\newcommand{\dpr}[1]{\textcolor{blue}{\{}\textcolor{blue}{\{}#1\textcolor{blue}{\}}\textcolor{blue}{\}}}
\newcommand{\mnb}[0]{... (More Neighbors) ...}
\usepackage[accepted]{icml2024}

\usepackage{amsmath}
\usepackage{amssymb}
\usepackage{mathtools}
\usepackage{amsthm}

\usepackage[capitalize,noabbrev]{cleveref}

\theoremstyle{plain}

\theoremstyle{definition}

\theoremstyle{remark}

\usepackage[textsize=tiny]{todonotes}

\icmltitlerunning{Similarity-based Neighbor Selection for Graph LLMs}

\begin{document}

\twocolumn[
\icmltitle{Similarity-based Neighbor Selection for Graph LLMs}



\icmlsetsymbol{equal}{*}

\begin{icmlauthorlist}
\icmlauthor{Rui Li}{yyy}
\icmlauthor{Jiwei Li}{comp}
\icmlauthor{Jiawei Han}{uiuc}
\icmlauthor{Guoyin Wang}{sch}
\end{icmlauthorlist}

\icmlaffiliation{yyy}{University of Science and Technology of China}
\icmlaffiliation{comp}{Zhejiang University}
\icmlaffiliation{sch}{Bytedance}
\icmlaffiliation{uiuc}{University of Illinois at Urbana-Champaign}
\icmlcorrespondingauthor{Rui Li}{rui\_li@mail.ustc.edu.cn}
\icmlcorrespondingauthor{Guoyin Wang}{guoyin.wang@bytedance.com}

\icmlkeywords{Machine Learning, ICML}

\vskip 0.3in
]



\printAffiliationsAndNotice{} 
\begin{abstract}
    Text-attributed graphs (TAGs) present unique challenges for direct processing by Language Learning Models (LLMs), yet their extensive commonsense knowledge and robust reasoning capabilities offer great promise for node classification in TAGs.
    Prior research in this field has grappled with issues such as over-squashing, heterophily, and ineffective graph information integration, further compounded by inconsistencies in dataset partitioning and underutilization of advanced LLMs.
    To address these challenges, we introduce Similarity-based Neighbor Selection (SNS).
    Using SimCSE and advanced neighbor selection techniques, SNS effectively improves the quality of selected neighbors, thereby improving graph representation and alleviating issues like over-squashing and heterophily. Besides, as an inductive and training-free approach, SNS demonstrates superior generalization and scalability over traditional GNN methods.  Our comprehensive experiments, adhering to standard dataset partitioning practices, demonstrate that SNS, through simple prompt interactions with LLMs,  consistently outperforms vanilla GNNs and achieves state-of-the-art results  on datasets like PubMed in node classification, showcasing LLMs' potential in graph structure understanding. Our research further underscores the significance of graph structure integration in LLM applications and identifies key factors for their success in node classification.
    Code is available at https://github.com/ruili33/SNS.
\end{abstract}

\section{Introduction}
Large Language Models (LLMs) 
\cite{openai2023gpt4,touvron2023llama,chowdhery2022palm}
 show remarkable potential in tackling reasoning-intensive tasks 
in a wide range of fields including 
language understanding \cite{openai2023gpt4}, multi-hop reasoning \cite{wei2022chain,fu2023chainofthought}, vision  \cite{wang2023visionllm,berrios2023towards}, and robotics \cite{zeng2023large,yu2023language}.  
Beyond these areas, much real-world data is structured in graph forms, which cannot be directly transformed into text. Text-attributed graphs (TAGs), characterized by nodes with textual attributes, represent a significant subset of such data.
In theory, TAGs' node classification task should 
significantly
benefit from LLMs: First,  its text-based nature could be dramatically enhanced by LLMs' ability on text understanding. Additionally, many node classification tasks require extensive commonsense knowledge, expert knowledge, and robust reasoning capabilities, which meet the expertise of LLMs. 
\begin{figure*}[t]
    \centering
    \begin{minipage}{1\textwidth}
        \centering
        \includegraphics[width=1\textwidth]{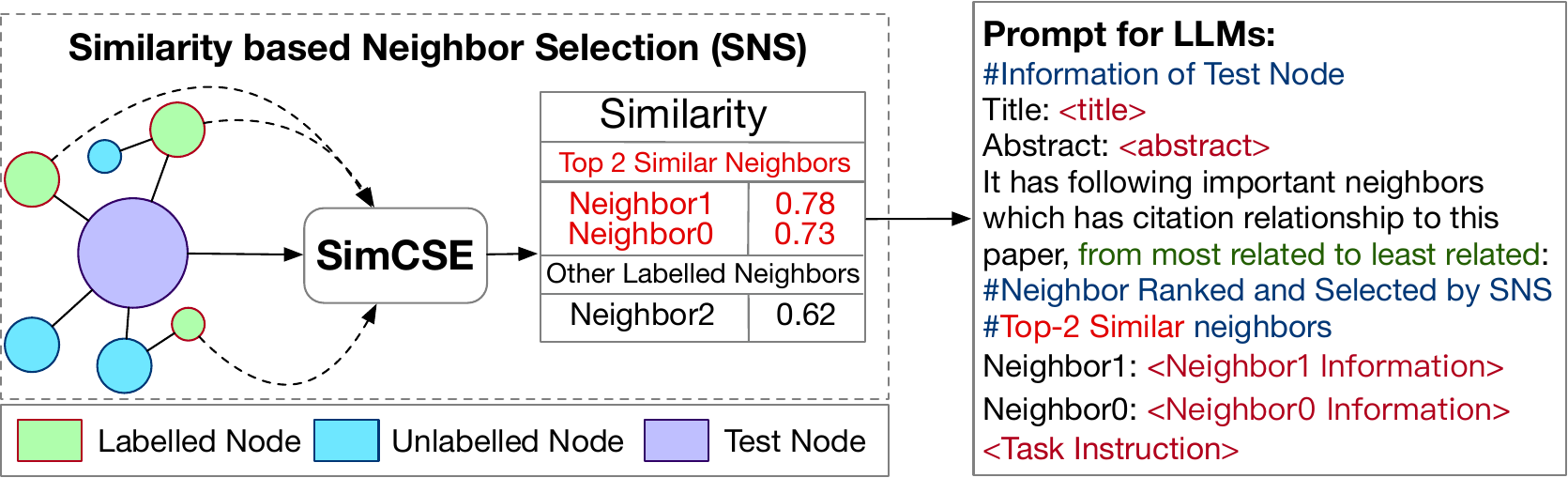}
    \end{minipage}
    \caption{An overview of SNS and the prompt for LLMs. {\color{red}Top 2 Similar Neighbors} are ranked and selected according to similarity determined by SimCSE \cite{gao2021simcse}, and subsequently incorporated into the prompts sequentially from most to least related.  }
    \label{fig:SNS}
\end{figure*}

Several recent studies have explored 
the path of
integrating graph structure information into prompts, including directly encoding node and edge lists \cite{liu2023evaluating,wang2023can}, summarization-based methods \cite{chen2023exploring,guo2023gpt4graph}, $\gamma$-hop random neighbor selection and attention based  methods \cite{huang2023llms}.
Unfortunately, as far as we are concerned, 
existing LLM-based node classification approaches still significantly underperform traditional
supervised graph learning models such as GCNs \cite{kipf2017semisupervised}  and GATs \cite{2018graph}. One key reason for the inferior performances of LLMs is the  inherent structure disparity between TAGs and natural language prompts.
Up to date, there hasn't been a consensus on how to effectively incorporate graph information into LLMs: 
$\gamma$-hop random neighbor selection (without any neighbor ranking) often leads to issues such as over-squashing \cite{topping2021understanding,deac2022expander,alon2020bottleneck} and heterophily; while attention based method  using LLMs for top-k neighbor selection is also shown to be 
 suboptimal \cite{gatto2023text} and problematic.
 Moreover, the partitioning of datasets for some existing research lacks consistency and standardization.

In response, 
we propose  similarity-based neighbor selection (SNS)   for  node classification  in TAGs. SNS begins with Recursive Neighbor Selection to discern labeled neighbors progressively from closer to more distant hops, tailored to each node's specific context. This is followed by Similarity-based Neighbor Ranking Strategy, drawing inspiration from the message-passing mechanism in GAT. Given the frequent occurrence of heterophily in graph structures, selecting appropriate neighbors is critical. We employ SimCSE \cite{gao2021simcse} to measure and rank the similarity between text attributes of nodes and their neighbors.  The top-ranking neighbors are then integrated into our prompts, improving graph information integration and mitigating issues like heterophily and over-squashing.
 Figure  \ref{fig:SNS} illustrates the SNS methodology and the prompt structure for LLMs.


 We conducted extensive experiments across 5 widely-used node classification datasets.
For a fair comparison with supervised GNN baselines, 
we adhered rigorously to the data partitioning strategies outlined in the graph community \cite{2018graph,hu2020open}. 
Throughout these experiments, we identified several critical factors that are pivotal for the success of LLMs in node classification:
\begin{enumerate}
    \item Recursive Neighbor Selection and Similarity-based Neighbor Ranking are crucial, as shown in Sections \ref{section:sns_method} and \ref{section:ablation}.
    \item The capabilities of LLMs matter. GPT-4 significantly outperforms GPT-3.5 in many datasets.
    \item Incorporating textual information from neighboring nodes generally enhances performance, though label inclusion's efficacy varies by dataset.
\end{enumerate}

SNS  consistently outperforms supervised graph learning baselines (GCN, GAT, and GraphSAGE \cite{hamilton2017inductive}) and all existing prompt-based methods across five widely-used benchmarks, achieving state-of-the-art results on PubMed.
To the best of our knowledge, SNS is the first prompt-based approach that 
 could achieve performance superior to vanilla GNNs on node classification.
 Meanwhile, LLMs significantly benefit from graph structure integration, showing marked improvements over vanilla zero-shot scenarios.
 
  Our extensive experiments reveal that, despite utilizing a linear training object, LLMs exhibit initial proficiency in handling node classification on graph-structured data. 
 Compared to GNN-based methods, SNS offers several advantages: it requires no training or fine-tuning, is inherently inductive and easily generalizes to new nodes and graphs, demonstrates better scalability to large graphs, and maintains constant time complexity for single-node prediction. 
  Building upon this observation, we highlight that with the extensive pre-trained knowledge base and advanced reasoning capabilities, LLMs hold the potential for generalizing to some tasks in  TAGs without the necessity for external tools or additional fine-tuning.

\begin{table*}[htbp]
    \caption{Results for five node classification datasets across  Zero-shot, Zero-shot CoT, Few-shot and Few-shot CoT using GPT-3.5.}
    \label{table:primi_exp}
    \begin{tabular}[]{>{\centering\arraybackslash} p{4.5cm} | >{\centering\arraybackslash} p{1.6cm}>{\centering\arraybackslash} p{1.6cm}>{\centering\arraybackslash} p{1.6cm}>{\centering\arraybackslash} p{2.5cm}>{\centering\arraybackslash} p{2.5cm}}
    \toprule

    Experiment Settings& Cora & PubMed   & CiteSeer & Ogbn-arxiv & Ogbn-products  \\
   \midrule

    Zero-shot        & 66.5  &89.2   &69.6   &73.4   &79.0       \\
    Zero-shot CoT &62.8&90.8&67.2&68.9&72.7\\
    Few-shot&66.9&88.2&66.0&71.5&80.8\\
    Few-shot CoT&65.4&86.4&64.7&69.4&73.8\\
    \bottomrule
\end{tabular}

\end{table*}

\section{Similarity based Neighbor Selection}

\subsection{Preliminaries}
\textbf{Text-Attributed Graphs:} A text-attributed graph (TAG) is a graph $G=(\mathcal{V},E)$, where $\mathcal{V}$ denotes the set of vertices and E denotes the set of edges. Each node $v\in V$ possesses an associated text attribute $T_v=\{w_1, ..., w_n\}$, where n denotes the length of the text sequence.\\
\textbf{Node Classification:} In a partially labeled graph, where only a subset of nodes $\mathcal{L} \in \mathcal{V}$ have labels $y_{\mathcal{L}}$, the objective is to deduce the labels $y_{U}$ for the unlabeled nodes $U=\mathcal{V}\backslash\mathcal{L}$, utilizing the graph's structure and the text attributes.
\subsection{Challenges}
\label{section:SNS_challenge}
Challenges current LLMs face for  the node classification task are four folds:

(1) {\bf Discerning neighbor importance: }
Directly incorporating information about randomly sampled 1-hop or 2-hop neighbors into prompts \cite{chen2023exploring,huang2023llms} without any ranking or selection has not yielded satisfactory results.
The reason behind this is that it is hard for LLMs  to
distinguish between varying levels of importance and relatedness 
when it comes to their neighbors. 

(2) {\bf Over-squashing}  \cite{topping2021understanding,deac2022expander,alon2020bottleneck}
denotes the situation where the model fails to leverage all available information 
when excessive neighbor information is supplied.
Over-squashing occurs when a large graph's information is compressed into a limited number of dimensions, resulting in information loss and potentially diminishing the effectiveness.
While prevalent in traditional graph learning methods like GCNs, over-squashing is particularly pronounced in LLMs when graph information is not adequately filtered, due to their intrinsic limited capacity for long-term dependencies \cite{liu2022relational} and the hard token limit on input length.

(3) {\bf Struggles with heterophilous graphs: }
Heterophily describes a situation where connected nodes in a graph are more likely to differ than resemble, a scenario contrasting with homophily.   For LLMs, given that their attention mechanisms fundamentally rely on similarity \cite{NIPS2017_3f5ee243}, heterophilous graphs present a significant challenge. 

(4) {\bf The dilemma of balancing increased information against the potential noise from distant neighbors.}
Previous research  often focuses on providing LLMs with nodes within $\gamma$-hop (e.g. $\gamma=2$), potentially leading to suboptimal graph information utilization. For nodes with abundant labeled 1-hop neighbors, this proximity may be sufficient for accurate LLM predictions, rendering the inclusion of 2-hop neighbors, which typically convey less relevance and more noise, unnecessary. In contrast, for nodes with sparse labeled 2-hop neighbors, expanding to more distant neighbors can reveal additional labeled nodes, thereby enhancing the decision-making data available to LLMs.

\subsection{Similarity based Neighbor Selection}
\label{section:sns_method}

To address the challenges above, we draw 
inspiration
from  GAT. GAT assigns varying weights to different neighbors based on attention scores through message passing. This selective attention enables GAT to focus on a subset of nodes, thereby alleviating the over-squashing issue. Additionally, this approach is advantageous in addressing the heterophily issue, as the model can assign lower weights to neighbors that exhibit heterophilous characteristics, allowing for more nuanced information aggregation.

To adapt this line of thinking to LLMs, we propose  Similarity based Neighbor Selection (SNS in short) shown in Figure \ref{fig:SNS}. SNS is composed of  two strategies:

\subsubsection{Recursive Neighbor Selection:} 
SNS commences with a Recursive Neighbor Selection process, beginning with  the retrieval of a node's 1-hop neighbors, followed by an exploration of successive neighboring hops. The search terminates upon identifying a sufficient number of labeled neighbors or reaching a pre-established maximum number of search layers $\gamma$ ( $\gamma=5$ in our experiments). The threshold number of neighbors, denoted as $\alpha$, is a layer-specific hyperparameter. In our experiments, we set $\alpha=2$ for the first layer, and $\alpha=1$ for the rest layers. 

The strategy of recursive neighbor selection    
resolves the challenge of incorporating distant neighbors. It allows for tailored decisions regarding neighbor selection for each node based on their specific context. For example, a node surrounded by numerous labeled neighbors may limit its search to a smaller hop distance. In contrast, nodes with fewer labeled neighbors may extend their search to more distant neighbors. This strategy effectively minimizes noise while ensuring adequate information is captured for accurate decision-making.
\begin{table*}[htbp]
  \caption{Prompts used in experiments across different settings. M represents the method to select neighbors, which could be SNS, $\gamma$-hop Random ($\gamma$-hop random neighbor selection) or 1-hop attention. The neighbors obtained by SNS are incorporated into the prompt in descending order of their cosine similarity scores. The \param{Neighbor Instruction for M} is shown in Table \ref{table:neighbor_instruction} in Appendix \ref{section:app_det_prompt}. The \param{Task Instruction w/o Neighbor} and \param{Task Instruction w/ Neighbor} for each dataset are shown in Table \ref{table:task_instruction_wo} and \ref{table:task_instruction_w} in Appendix \ref{section:app_det_prompt}.}
  \label{table:prompt}
    \begin{tabular}[]{>{\centering\arraybackslash} p{3.05cm}  >{\centering\arraybackslash} p{13.25cm}}
    \toprule

    Experiment Settings& Prompts\\
   
   \midrule

     Vanilla Zero-shot        &   Title: \param{title}\nl Abstract: \param{abstract}\nl  \param{Task Instruction w/o Neighbor} \\
    \midrule
    M+label &    Title: \param{title}\nl Abstract: \param{abstract}\nl \param{Neighbor Instruction for M}\nl Neighbor Paper0: \dpr{Category: \param{Neighbor Paper0 Label}}\nl\nl \mnb \nl \param{Task Instruction w/ Neighbor}    \\
    \midrule
    M+text&    Title: \param{title}\nl Abstract: \param{abstract}\nl \param{Neighbor Instruction for M}\nl Neighbor Paper0: \dpr{Title: \param{Neighbor Paper0 Title}}\nl\nl \mnb \nl \param{Task Instruction w/ Neighbor}    \\
    \midrule
    M+text+lebel &  Title: \param{title}\nl Abstract: \param{abstract}\nl \param{Neighbor Instruction for M}\nl Neighbor Paper0: \dpr{Category: \param{Neighbor Paper0 Label}\nl Title: \param{Neighbor Paper0 Title}}\nl\nl \mnb \nl \param{Task Instruction w/ Neighbor}     \\
 
    \bottomrule
\end{tabular}

\end{table*}
    
\subsubsection{Similarity-based Neighbor Ranking Strategy:} 
SNS proceeds with a Similarity-based Neighbor Ranking Strategy, selectively integrating neighbor nodes into the prompt, prioritizing those with higher similarity to the target node's attributes.  We adopt SimCSE \cite{gao2021simcse} as our similarity metric, calculating the cosine similarity between the text attribute of each node and that of its neighbors.  Neighbors are then ranked based on this similarity score. Subsequently, we select the top-k neighbors (where k is a hyperparameter, shown in Table \ref{table:exp_det_k}) demonstrating the highest similarity. Upon identifying the top-ranking neighbors, they are sequentially added to the prompt, arranged by descending order of their cosine similarity scores.

The integration of SimCSE into our Similarity-based Neighbor Ranking Strategy adeptly addresses the difficulty LLMs face in evaluating neighbor importance. By prioritizing neighbors with high similarity scores, this method efficiently filters out lower-quality graph information, optimizing the use of limited context capacity and mitigating over-squashing. Additionally, it skillfully addresses heterophily by excluding dissimilar neighbors, as the similarity metric inherently selects nodes with analogous characteristics, thus elevating the relevance and quality of integrated neighbor data.

\subsection{Discussion}
SNS exhibits several advantageous properties compared to 
traditional GNN-based methods:

(1) First, SNS offers a zero-shot solution for node classification with LLMs, delivering promising performance without the need for training or fine-tuning.

(2) Second,  as a prompt-based approach, SNS is inherently inductive, facilitating straightforward generalization to new nodes and graphs.

(3) Third,  the method boasts superior scalability, with a constant time complexity for single-node processing, in contrast to the significant increase in time complexity and memory usage experienced by GNNs in larger graph applications.

(4) Finally, SNS effectively leverages the extensive commonsense knowledge and robust reasoning capabilities of cutting-edge LLMs.

\section{Experiments}

\subsection{Datasets}
In this study, we evaluate SNS and baselines on five widely-used node classification benchmarks: 
Cora \cite{mccallum2000automating}, PubMed \cite{sen2008collective}, CiteSeer \cite{giles1998citeseer}, Ogbn-arxiv and Ogbn-products \cite{hu2020open}. To enable apple-to-apple  comparisons, our experimental setup adheres closely to the methodologies outlined in \citet{2018graph,hu2020open}. Specifically, for Cora, PubMed and CiteSeer, our experimental design includes 20 training examples per class, alongside a validation set of 500 nodes and a test set of 1,000 nodes. For Ogbn-arxiv and Ogbn-products, we  maintain the original dataset partitions as established in \citet{hu2020open}. Considering the high cost associated with the OpenAI API, we randomly sampled a subset of 1,000 nodes from the test sets of Ogbn-arxiv and Ogbn-products for our evaluation.

\begin{table*}[htbp]
    \caption{Comparison of  SNS with $\gamma$-hop random neighbor selection ($\gamma$-hop Random in the table, $\gamma=1, 2, 3$), 1-hop attention \cite{huang2023llms}, vanilla zero-shot  in text+label scenarios, alongside SNS performance across text-only, label-only, and combined text and label scenarios.
    Considering the memory issue, the GAT results on Ogbn-products are derived through GAT with NeighborSampling \cite{2018graph,hamilton2017inductive}.}
\label{table:main_results}
    \begin{tabular}[]{>{\centering\arraybackslash} p{5.2cm} | >{\centering\arraybackslash} p{1.58cm}>{\centering\arraybackslash} p{1.5cm}>{\centering\arraybackslash} p{1.5cm}>{\centering\arraybackslash} p{2.5cm}>{\centering\arraybackslash} p{2.5cm}}
    \toprule

    Experiment Settings& Cora & PubMed   & CiteSeer & Ogbn-arxiv & Ogbn-products  \\
    \midrule
   &\multicolumn{5}{c}{\textit{GPT3.5}} \\
   \midrule

    Vanilla Zero-shot        & 66.5  &89.5   &69.6   &73.4   &79.0       \\
    1-hop Attention+text+lebel  & 72.4  &90.6   & 71.5  &72.3   &   82.0    \\
    1-hop Random+text+lebel&72.2 &90.7&71.6&72.7&82.5\\
    2-hop Random+text+lebel&75.6 &91.0&72.9&71.6&82.0\\
    3-hop Random+text+lebel& 70.5&90.8&73.0&72.6&80.4\\
    \midrule
    SNS+text+lebel  &78.5   &91.4   &\textbf{75.0}   &\textbf{74.4}   &84.6\\
    SNS+label &78.5&90.6  &73.6 &\textbf{74.4}  &83.8\\
    SNS+text &69.0&92.1&71.4&73.7&80.5\\

    \midrule
    &\multicolumn{5}{c}{\textit{GPT4}} \\
    \midrule

    Vanilla Zero-shot        & 67.4  &92.8   &72.3   &72.9   &80.1       \\

    SNS+text+lebel  &\textbf{82.5}   &\textbf{93.8}   &74.6   &74.2   &\textbf{88.3}\\
    \midrule
    &\multicolumn{5}{c}{\textit{GNNs}} \\
    \midrule
    GCN~\cite{kipf2017semisupervised}&79.3 $\pm$ 0.2 & $79.9\pm 0.2$& $72.1\pm 0.5$   &$70.1\pm0.3$   &$76.2\pm 0.3$\\
    GAT~\cite{2018graph}         &81.4 $\pm$ 0.4  &$79.9\pm 0.9$   &$72.3\pm 0.1$   &71.0 $\pm$ 0.3   &$81.3\pm 0.3$*\\
    GraphSAGE~\cite{hamilton2017inductive} &80.4 $\pm$ 0.5 & $79.3\pm 0.2$& $73.4\pm 0.1$   &$71.3\pm0.2$   &$78.4\pm 0.4$\\

    \bottomrule
\end{tabular}

\end{table*}

\subsection{Preliminary Experiments}
\label{section:exp_primin_exp}
To identify the most suitable prompting methods for node classification, we conducted trials on all 5 datasets using mainstream approaches: Zero-shot, Zero-shot with Chain-of-Thought (Zero-shot CoT)  \cite{kojima2022large}, Few-shot \cite{brown2020language}, Few-shot CoT \cite{wei2022chain}, employing GPT-3.5. The results are shown in Table \ref{table:primi_exp}.

Our findings align with  \citet{huang2023llms,wei2022chain}, reveal limited or negative benefits of few-shot and CoT prompting on node classification tasks. Few-shot prompting's limited effectiveness is due to the node classification task's complexity, which demands more specialized knowledge compared to standard text classification tasks. For example, Ogbn-arxiv involves classifying 40 subfields in computer science, which  requires a breadth of knowledge beyond  the boundary of few-shot learning. Another possible explanation comes from \citet{li2023human}, which suggest that LLMs can attain few-shot performance levels even in zero-shot settings, highlighting LLM's zero-shot ability. CoT techniques, meanwhile, prove less effective due to the knowledge-driven rather than reasoning-driven nature of node classification, potentially introducing extraneous noise. Hence, we employed zero-shot prompting in our primary experiments for its relative effectiveness and simplicity.

\subsection{Experimental Settings and Baselines}
We compare SNS against three baselines: $\gamma$-hop random neighbor selection ($\gamma$-hop Random in short, $\gamma=1, 2, 3$), 1-hop attention\footnote{Limited by LLMs' context capacity, attention-based methods operate within a 1-hop range, consistent with \citet{huang2023llms}.}  \cite{huang2023llms} and vanilla zero-shot scenario. 
$\gamma$-hop random neighbor selection involves integrating data from randomly selected neighbors within a  $\gamma$-hop radius (aka the 1-hop and 2-hop prompting in \citet{huang2023llms}), while vanilla zero-shot scenario does not incorporate any graph information.
As mentioned in Section \ref{section:exp_primin_exp}, our evaluation focuses on zero-shot prompting.
Additionally, we assessed SNS against three baselines under the text+label scenario and explored various strategies for incorporating neighbor information into SNS, namely text-only, label-only, and text+label approaches.
  The prompts used in our main experiments are shown in Table 
\ref{table:prompt}. For comparison with traditional graph-based methods, we utilize  GCN \cite{kipf2017semisupervised}, GAT \cite{2018graph} and GraphSAGE \cite{hamilton2017inductive} as our graph baselines.

In our main experiments, we utilize GPT3.5 (gpt-3.5-turbo), GPT4  \citep{openai2023gpt4}\footnote{https://platform.openai.com/docs/models} as the model backbone.  GPT-3.5 serves as the default model for our ablation studies, unless stated otherwise.

\subsection{Results}

The results summarized in Table \ref{table:main_results} demonstrate that the integration of graph structures and neighbor information leads to LLMs consistently outperforming Vanilla GNNs by considerable margins in zero-shot  scenarios, achieving state-of-the-art performance on PubMed.  Moreover, across all datasets, LLMs show enhanced performance when including SNS-augmented graph information within the prompts, significantly exceeding the results in vanilla zero-shot scenarios. These findings underscore the potential of LLMs, with suitably crafted prompts and judicious neighbor selection, to effectively utilize graph information and exhibit emerging capabilities in node classification tasks within TAGs.

On the other hand, the consistent superiority of LLMs using SNS over $\gamma$-hop random neighbor selection and 1-hop attention is evident.  This demonstrates the superiority of SNS  in enhancing textual graph representation and graph information utilization.

Notably, in  PubMed, the incorporation of neighbor information remains efficacious even at an advanced stage of model performance, enhancing accuracy from 92.8\% in the vanilla zero-shot scenario to 93.8\%, thereby achieving  state-of-the-art results\footnote{The previous state-of-the-art method on PubMed, ACM-Snowball-3 \cite{luan2021heterophily}, achieved an accuracy of 91.44\% (is surpassed by our performance, 93.8\%), as reported on https://paperswithcode.com/sota/node-classification-on-pubmed.}.  This observation compellingly suggests that, in many scenarios, graph information retains its utility for LLMs even after achieving substantial initial success.

Furthermore, our results consistently demonstrate that incorporating textual information into the prompts enhances performance in all scenarios for GPT3.5. This aligns with our intuition that neighbors' textual attributes offer valuable context about the subject paper and its neighbors. However, the impact of integrating label information varies across datasets. It significantly enhances performance in Cora and Ogbn-products, but leads to a decline in PubMed. This disparity likely stems from differences in top-k neighbor accuracy among datasets, as illustrated in Figure \ref{fig:quality_simcse}.

\begin{figure}[t]
    \centering
    \begin{minipage}{0.48\textwidth}
        \centering
        \includegraphics[width=1\textwidth]{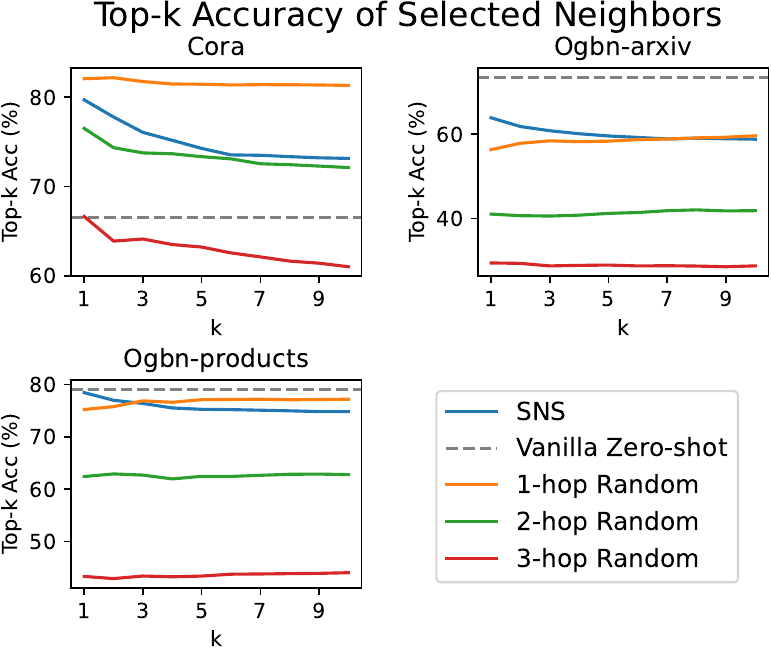}
    \end{minipage}

    \caption{Top-k neighbors accuracy of SNS, random neighbor selection, and LLMs' vanilla zero-shot accuracy  across 3 datasets.  }
    \label{fig:quality}
\end{figure}
\begin{table}[t]
    \caption{Failure (failed to find labeled neighbors) rate of each selection method in Cora. $\gamma$-hop Ran represents $\gamma$-hop random neighbor selection.}
    \label{table:abla_cora_neighbor}
    \begin{tabular}[]{>{\centering\arraybackslash} p{1.1cm} | >{\centering\arraybackslash} p{1.44cm}>{\centering\arraybackslash} p{1.44cm}>{\centering\arraybackslash} p{1.44cm}>{\centering\arraybackslash} p{0.8cm}}
    \toprule
    & 1-hop Ran& 2-hop Ran&3-hop Ran&SNS  \\
    \midrule
    Cora &40.3&10.6&5.3&\textbf{4.9}\\
    \bottomrule
\end{tabular}

\end{table}

\section{Ablation Studies}
\label{section:ablation}
\textbf{Quality of Neighbors between SNS and Random Neighbor Selection.} To evaluate the neighbor selection quality of SNS and $\gamma$-hop random neighbor selection, we investigate top-k neighbor accuracy, reflecting the proportion of top-k neighbors sharing the same label with the test node, as shown in Figure \ref{fig:quality}. 
Additionally, Table \ref{table:abla_cora_neighbor} presents the failure rates for each method in Cora, indicating the frequency at which these methods fail to identify a labeled neighbor.

Despite including distant neighbors and greatly enriching graph information (shown in Table \ref{table:abla_cora_neighbor}), SNS maintains competitive top-k accuracy relative to 1-hop random selection, underscoring its superior neighbor quality. Besides, SNS effectively mitigates the issue of heterophily in distant neighbors, as evidenced by its superior top-k neighbor accuracy compared to 2-hop and 3-hop random neighbor selection. 


Furthermore,  the observation that the zero-shot accuracy significantly exceeds the top-k neighbor accuracy  in  Ogbn-arxiv and PubMed (see Figure \ref{fig:quality_simcse}) might potentially explain why there is little increase in accuracy, or even a decrease, after adding the labels of neighbors. 

\textbf{SimCSE: } Comparative analysis of SNS performance with and without the ranking of SimCSE \cite{gao2021simcse}, as presented in Table \ref*{table:simcse}. Besides, the top-k accuracy of neighbors selected by SNS with and without ranking by SimCSE are shown in Figure \ref{fig:quality_simcse} in Appendix \ref{section:app_det_abla_simcse}.  The findings, encompassing both overall performance and top-k neighbor accuracy, consistently demonstrate the enhanced efficacy of SNS with SimCSE, highlighting the critical role of SimCSE in neighbor ranking.
\begin{table*}[htbp]
    \caption{Comparison of SNS performance with and without SimCSE in a zero-shot scenario using GPT-3.5.}
    \label{table:simcse}
    \begin{tabular}[]{>{\centering\arraybackslash} p{5.2cm} | >{\centering\arraybackslash} p{1.58cm}>{\centering\arraybackslash} p{1.5cm}>{\centering\arraybackslash} p{1.5cm}>{\centering\arraybackslash} p{2.5cm}>{\centering\arraybackslash} p{2.5cm}}
    \toprule

    Experiment Settings& Cora & PubMed   & CiteSeer & Ogbn-arxiv & Ogbn-products  \\
    \midrule
    
    SNS w/o SimCSE+label&77.2           &90.4&73.6&73.6&82.4\\
    SNS w/ SimCSE+label &\textbf{78.5}  &90.6  &73.6 &\textbf{74.4}  &83.8\\
    SNS w/o SimCSE+text &68.8           &91.1&70.0&73.3&79.5\\
    SNS w/ SimCSE+text &69.0               &\textbf{92.1}&71.4&73.7&80.5\\
    SNS w/o SimCSE+text+lebel&77.8      &90.8&74.1&73.8&82.7\\
SNS w/ SimCSE+text+lebel&\textbf{78.5}   &91.4   &\textbf{75.0}   &\textbf{74.4}   &\textbf{84.6}\\

    \bottomrule
\end{tabular}

\end{table*}

\begin{figure}[t]
    \centering
    \begin{minipage}{0.5\textwidth}
        \centering
        \includegraphics[width=1\textwidth]{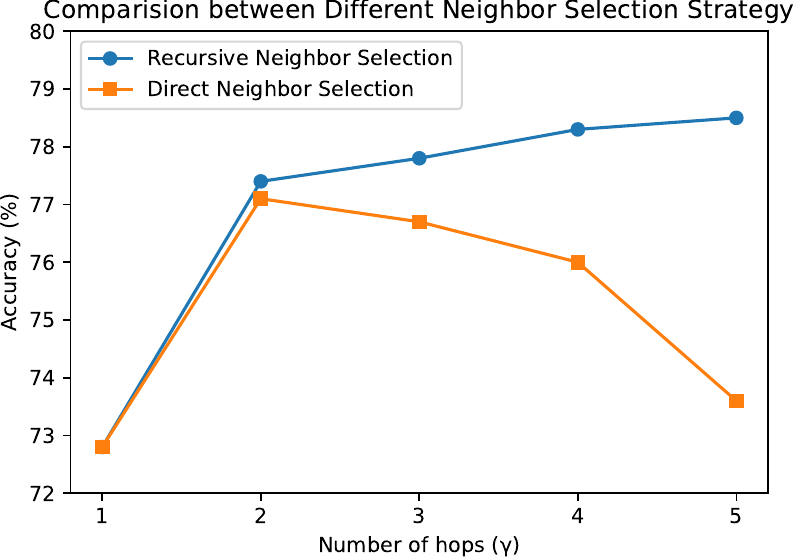}
    \end{minipage}
    
    \caption{Comparison between Recursive Neighbor Selection and Direct Selection across different hops ($\gamma$).}
    \label{fig:direct}
\end{figure}
    
\textbf{Recursive Neighbor Selection vs Direct Selection.} We contrast Recursive Neighbor Selection with Direct Selection, the latter being a simple method in $\gamma$-hop random neighbor selection \cite{chen2023exploring,huang2023llms}, characterized by the inclusion of all labeled neighbors within $\gamma$-hop, without customizing for each node. This comparative analysis focuses on their respective performances across varying hop counts (denoted as $\gamma$). After selecting neighbors, we apply Similarity-based Neighbor Ranking Strategy for both methods.  The results on cora are delineated in Figure \ref{fig:direct}.

The findings reveal that Recursive Neighbor Selection significantly surpasses Direct Selection in efficacy, particularly as the number of hops ($\gamma$) increases. Notably, the performance of Recursive Neighbor Selection improves with larger values of $\gamma$, suggesting its effective utilization of distant neighbors' information. Conversely, Direct Selection demonstrates a decline in performance with increased hop counts, indicative of an over-squashing problem inherent to this approach.

\begin{figure}[t]
    \centering
    \begin{minipage}{0.46\textwidth}
        \centering
        \includegraphics[width=0.95\textwidth]{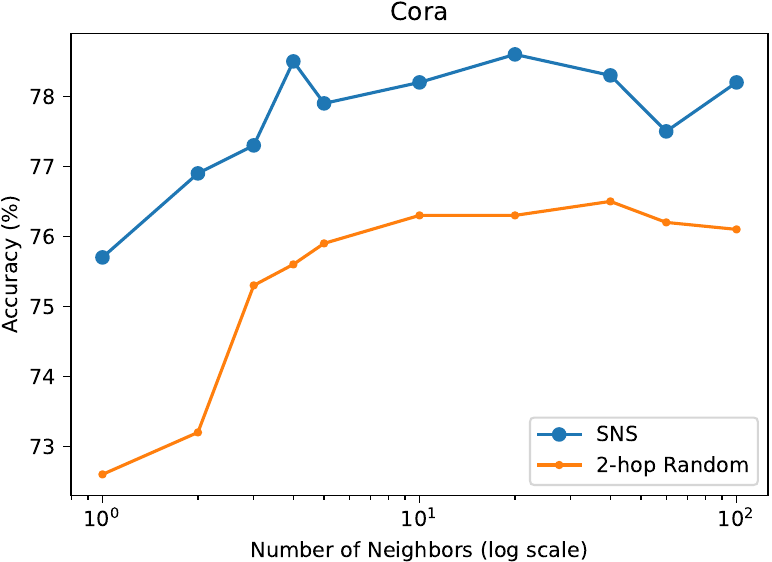}
    \end{minipage}
    \caption{The results of SNS and 2-hop random neighbor selection for Cora across different numbers of neighbors (k).}
    \label{fig:neighbors_cora}
\end{figure}

\textbf{Number of Neighbors} We examined the impact of varying neighbor count (parameter k in Section \ref{section:sns_method}) on the performance of SNS and 2-hop random neighbor selection   for Cora, as depicted in Figure \ref{fig:neighbors_cora}.
The results reveal that LLMs benefit from an increased number of neighbors when the initial count is low. However, performance reaches a plateau beyond a certain threshold, with minor fluctuations. This finding aligns with our discussion on over-squashing in LLMs in Section \ref{section:SNS_challenge}, highlighting LLMs' limited utility from excessive neighbor information at high k values, resulting in performance stagnation.


\begin{table}[htbp]
    \caption{Comparison between TAPE (on GCN, GraphSAGE, RevGAT \cite{li2021training}) and SNS.}
    \label{table:abla_comp_tape}
    \begin{tabular}[]{>{\centering\arraybackslash} p{2.6cm} | >{\centering\arraybackslash} p{1.98cm}>{\centering\arraybackslash} p{2.3cm}}
    \toprule

    & Cora & PubMed  \\
    \midrule
    TAPE+GCN           & 79.0  $\pm$  0.7  &81.8  $\pm$  1.0 \\
    TAPE+SAGE          & 79.2  $\pm$  0.4  & 86.1  $\pm$ 2.1  \\
    TAPE+RevGAT        & 78.4 $\pm$  1.2 & 85.9 $\pm$ 1.9 \\
    \midrule
    SNS-GPT3.5 &78.5&91.4\\
    SNS-GPT4 &\textbf{82.5}&\textbf{93.8}\\
    \bottomrule
\end{tabular}

\end{table}

\begin{figure}[t]
    \centering
    \begin{minipage}{0.45\textwidth}
        \centering
        \includegraphics[width=0.95\textwidth]{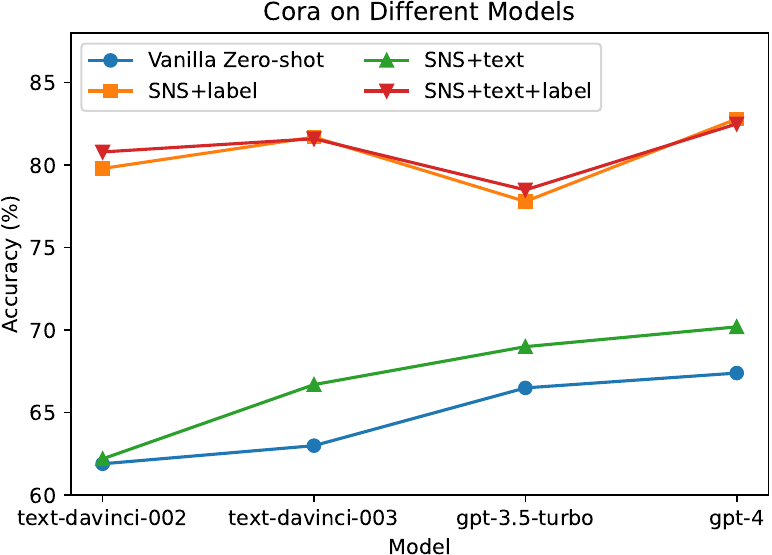}
    \end{minipage}
    \caption{The performance of SNS across LLMs with varying capabilities on Cora.}
    \label{fig:models_cora}
\end{figure}

\textbf{Comparison between SNS with TAPE \cite{he2023harnessing}.} To elucidate the differential roles of LLMs in enhancement and prediction, we conducted a comparative analysis of SNS and TAPE. It is important to note that due to the differing dataset splits from our study  to \citet{he2023harnessing}, the results of TAPE vary from their study. The comparative results are detailed in Table \ref{table:abla_comp_tape}.

Our analysis reveals that when LLMs are employed primarily for prediction, as in our method, they are more adept  at addressing challenges in low labeling settings where insufficient data is available for fine-tuning Pre-trained Language Models (PLMs).  
Furthermore, our approach demonstrates superior scalability compared to the method employed by \citet{he2023harnessing}. Their high-cost approach for generating LLM explanations led to challenges in processing the ogbn-products dataset, necessitating subgraph sampling. In contrast, our method scales more efficiently to larger datasets, maintaining a computational complexity of O(1) for single-node prediction.

\textbf{SNS on Different Models: } Figure \ref{fig:models_cora} and \ref{fig:models_prod} display SNS's performance across four notable LLMs, ranging from less to more advanced models, including text-davinci-002, text-davinci-003, gpt-3.5-turbo, and gpt-4, evaluated on Cora and Ogbn-products datasets. The results strongly indicate SNS's efficacy across various models, underscoring the importance of graph information. Further details and discussions are provided in Appendix \ref{section:app_models}.

\section{Related Work}


\subsection{GNNs for TAGs}
GNNs have been developed to process graph data, building upon the message-passing framework \cite{gilmer2017neural}. This has led to the creation of various GNN variants. GCN \cite{kipf2017semisupervised} adapts convolutional structures to graph data, while GAT \cite{2018graph} leverages the attention mechanism for enhanced message-passing expressiveness \cite{liu2023towards}. GraphSAGE \cite{hamilton2017inductive} provides a scalable inductive approach for generating embeddings for new data. However, some of these methods often require training on entire graphs, potentially limiting scalability and generalization to new graphs.


\subsection{PLMs for TAGs} PLMs \cite{devlin2018bert,lewis-etal-2020-bart,radford2019language} could further enhance GNNs by enriching node embeddings. One notable methodology is the  cascaded architecture \cite{zhu2021textgnn,chien2021node,hu2020gpt,duan2023simteg}, where the generation of node embeddings is independent from the training of GNNs. Besides, some works belong to iterative architecture, such as DRAGON \cite{yasunaga2022deep}, Graphormer \cite{yang2021graphformers}  and Heterformer \cite{jin2023heterformer}. This architecture trains PLMs and GNNs in an iterative strategy, enabling PLMs to get insights from the graph structure. 
Since these studies are initially designed for the link-prediction task, which lacks solid semantic interpretation, they are not included as our baselines.

\subsection{LLMs for TAGs}
The integration of LLMs and graph data has received increasing attention across the community.  InstructGLM \cite{ye2023natural} and GraphGPT \cite{tang2023graphgpt} adopted instruction tuning to apply  LLMs to TAGs. LLMtoGraph \cite{liu2023evaluating} and NLGraph \cite{wang2023can} integrate graph structures by crafting prompts containing node and edge lists. 
GraphText \cite{zhao2023graphtext} generates graph-syntax trees to encapsulate knowledge in the graph structure. GPT4Graph \cite{guo2023gpt4graph} uses LLMs with prompts that include context summarization and format explanation.  Another study, examining LLMs as Enhancers and Predictors, reveals initial efficacy in incorporating neighbors' text summarization \cite{chen2023exploring}. 

\citet{huang2023llms} introduce 1-hop and 2-hop random neighbor selection as well as  1-hop attention prompting, positing that structural data can improve LLM performance, especially with limited textual node attributes. However, their results show 1-hop attention often underperforms compared to 1-hop and 2-hop random selection. Their approach also lacks a thorough integration of distant neighbors and uses LLMs for neighbor ranking, a method critiqued for inefficiency \cite{gatto2023text} and problematic given LLMs' context limitations.

Consequently, these studies fall short of optimally leveraging graph information. Furthermore, some previous research, including \citet{chen2023exploring,huang2023llms}, due to not following the same training and testing set partition as \citet{2018graph}, their results are not directly comparable with those of GNNs and ours.
To the best of our knowledge, our method is the first prompt-based approach to consistently outperform the simplest GNNs, namely GCN, GAT and GraphSAGE. 

Besides, there is also a line of work that utilizes LLMs to strengthen existing graph learning frameworks, such as enhancing node features \cite{he2023harnessing,chen2023exploring}. \citet{liu2023one} employ LLMs to standardize node features across diverse domains. Also, some previous study also tries to utilize external tools, such as Iterative Reading-then-Reasoning in StructGPT \cite{jiang2023structgpt} and Graph-ToolFormer \cite{zhang2023graphtoolformer}.

\section{Conclusion}
In this paper, we present Similarity based Neighbor Selection (SNS), a straightforward yet versatile and efficacious high-quality neighbor filtration technique for prompt-based solutions to node classification. SNS offers a potential  mitigation to challenges such as over-squashing and diminished performance on heterophilous graphs.

Our comprehensive experimental analysis across diverse datasets and neighbor integration methods 
show that SNS not only markedly outperforms existing prompt-based approaches, but also surpasses Vanilla GNNs in zero-shot predictions, and even achieves state-of-the-art results on PubMed.
 A fine-grained analysis further  underscores the importance of graph integration and the effectiveness of each component of the SNS method, and particularly highlights SNS's effectiveness in sparse labeling scenarios. These results illuminate the potential of LLMs, when equipped with carefully designed prompts and strategic neighbor selection, to proficiently harness graph information, thereby exhibiting advanced capabilities in node classification tasks within TAGs.

\section*{Impact Statement}
This paper presents work whose goal is to advance the application of LLMs on the graph area. There are many potential societal consequences of our work, none which we feel must be specifically highlighted here.

\bibliography{example_paper}
\bibliographystyle{icml2024}

\newpage
\appendix
\onecolumn
\section{SNS on Different Models}
\label{section:app_models}
\begin{figure*}[t]
    \centering
    \begin{minipage}{0.46\textwidth}
        \centering
        \includegraphics[width=0.95\linewidth]{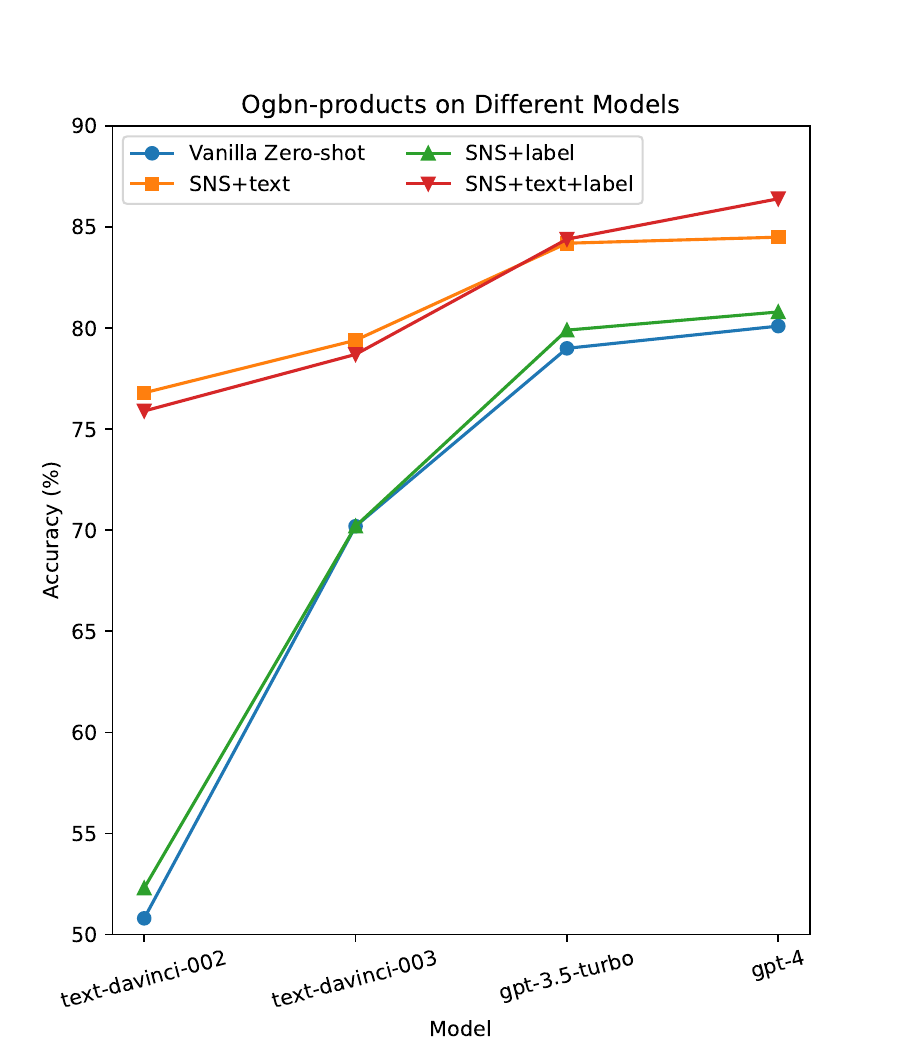}
    \end{minipage}  
    \caption{The performance of SNS across LLMs with varying capabilities on Ogbn-products.}
    \label{fig:models_prod}
\end{figure*}

 To further evestigate the performance of SNS across LLMs with varying capabilities, we conducted experiments utilizing four renowned LLMs, ranging from less powerful to more advanced models. These include text-davinci-002, text-davinci-003, gpt-3.5-turbo, and gpt-4. The datasets employed for this study were Cora and Ogbn-products\footnote{To accommodate the token limitations of the text-davinci-002 and text-davinci-003 models in the Ogbn-products dataset, our experimental setup was adjusted to a setting of k=1.}. The outcomes of these experiments are presented in Figure \ref{fig:models_cora} and \ref{fig:models_prod}.

In our analysis of both datasets, it is evident that all four models significantly benefit from the incorporation of graph information. Notably, in the case of the Cora dataset, the davinci models exhibit better performance compared to GPT-3.5 when graph information is added. This is in contrast to the zero-shot scenario, where GPT-3.5 surpasses the davinci models. Conversely, in the Ogbn-products dataset, GPT-3.5 consistently outperforms the davinci models.

\section{Details of Ablation Study}
\subsection{SimCSE}
\label{section:app_det_abla_simcse}
\begin{figure}[h]
    \centering
    \begin{minipage}{0.48\textwidth}
        \centering
        \includegraphics[width=1\textwidth]{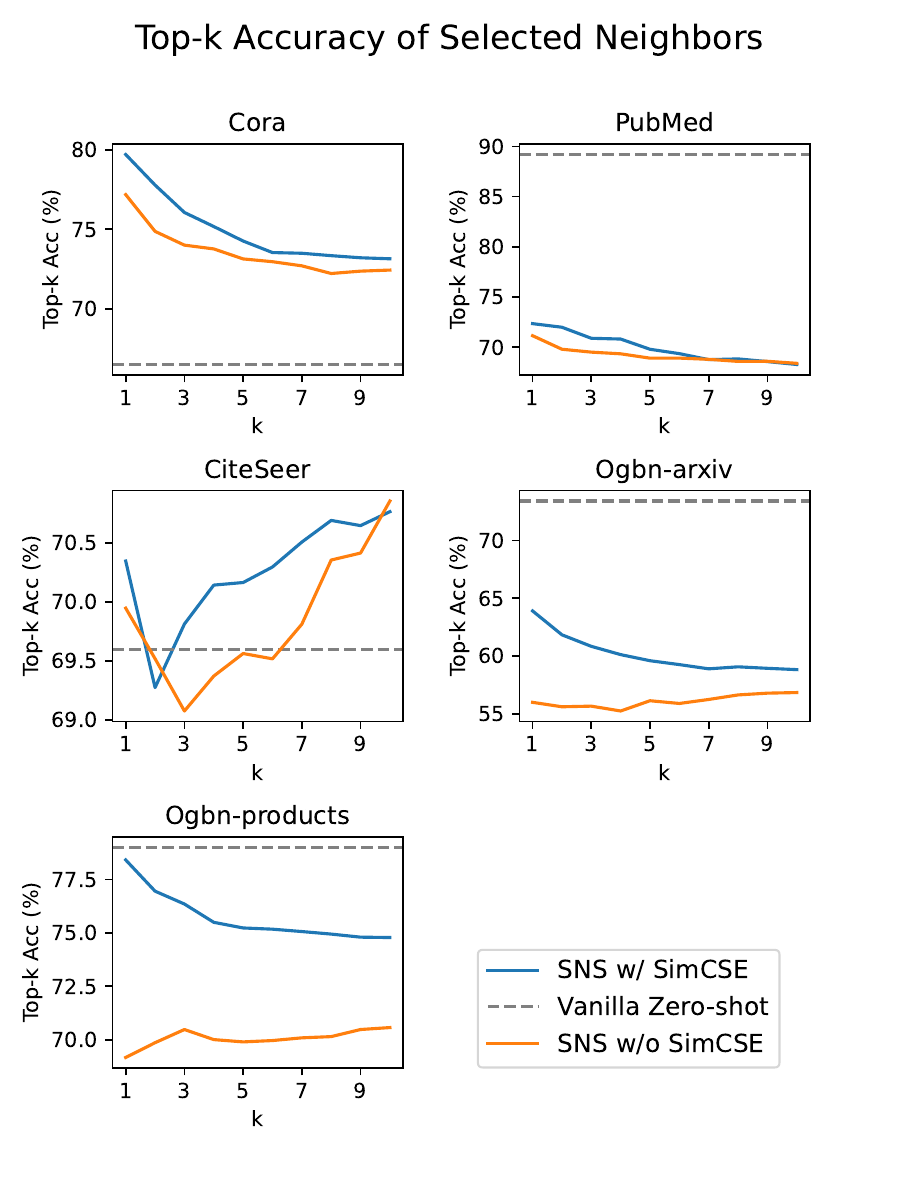}
    \end{minipage}
    \caption{Top-k neighbors accuracy of SNS with and without SimCSE as well as the vanilla zero-shot performance of LLMs across  all 5 datasets.  }
    \label{fig:quality_simcse}
\end{figure}

the top-k accuracy of neighbors selected by SNS with and without ranking by SimCSE are shown in Figure \ref{fig:quality_simcse}.

\section{Details of Main Experiments}
\subsection{Details of Experimental Setup}
The upper limit of the number of neighbors selected for each dataset (k) are shown in Table \ref{table:exp_det_k}.

\begin{table*}[htbp]
    \caption{The upper limit of the number of neighbors  selected for each dataset (k).}
    \label{table:exp_det_k}
    \begin{tabular}[]{>{\centering\arraybackslash} p{4.5cm} | >{\centering\arraybackslash} p{1.6cm}>{\centering\arraybackslash} p{1.6cm}>{\centering\arraybackslash} p{1.6cm}>{\centering\arraybackslash} p{2.5cm}>{\centering\arraybackslash} p{2.5cm}}
    \toprule

    & Cora & PubMed   & CiteSeer & Ogbn-arxiv & Ogbn-products  \\
    \midrule
    number of neighbors (k)            &4  &4 &8  &4   &100\\

    \bottomrule
\end{tabular}

\end{table*}

\subsection{Details of Prompts Design}
\label{section:app_det_prompt}
The prompts used in the main experiments are as shown in the Table \ref{table:prompt}. In the table, M represents the method to select neighbors, could be SNS, $\gamma$-hop Random ($\gamma$-hop random neighbor selection) or 1-hop attention. The \param{Neighbor Instruction for M} for SNS, 1-hop attention and $\gamma$-hop Random are shown in Table \ref{table:neighbor_instruction}.  The \param{Task Instruction w/o Neighbor} is shown in Table \ref{table:task_instruction_wo}. The \param{Task Instruction w/ Neighbor} is shown in Table \ref{table:task_instruction_w}. 
 Some ideas in designing our prompts are inspired by \citet{huang2023llms,chen2023exploring,he2023harnessing}

\begin{table*}[htbp]
    \caption{The \param{Neighbor Instruction for M} for SNS, 1-hop attention and $\gamma$-hop Random in Table \ref{table:prompt}.}
    \label{table:neighbor_instruction}
    \begin{tabular}[]{>{\centering\arraybackslash} p{2.8cm} | >{\centering\arraybackslash} p{7cm}>{\centering\arraybackslash} p{6cm}}
    \toprule

    M& SNS \& 1-hop attention & $\gamma$-hop Ran   \\
    \midrule
    \param{Neighbor Instruction for M}          &It has following important neighbors which has citation relationship to this paper, from most related to least related:  &It has following important neighbors which has citation relationship to this paper: \\

    \bottomrule
\end{tabular}

\end{table*}

\begin{table*}[htbp]
    \caption{The \param{Task Instruction w/o Neighbor} in Table \ref{table:prompt}.}
    \label{table:task_instruction_wo}
    \begin{tabular}[]{>{\centering\arraybackslash} p{3cm} | >{\centering\arraybackslash} p{13.2cm}}
    \toprule

    & \param{Task Instruction w/o Neighbor}  \\
    \midrule
    Cora  & Task: \nl There are following categories: \nl ['Rule Learning', 'Case Based', 'Genetic Algorithms', 'Theory', 'Reinforcement Learning', 'Probabilistic Methods', 'Neural Networks']\nl Which category does this paper belong to?\nl Output the most 1 possible category of this paper as a python list, like ['XX']\\
    \midrule
    PubMed& Question: Does the paper involve any cases of ['Type 1 diabetes'], ['Type 2 diabetes'], or ['Experimentally induced diabetes']? Output the most 1 possible category of this paper as a python list and in the form Category: ['XX'].\\
    \midrule
    CiteSeer &Task: \nl There are following categories: \nl ['Agents', 'Machine Learning', 'Information Retrieval', 'Database', 'Human Computer Interaction', 'Artificial Intelligence']\nl Which category does this paper belong to?\nl Output the most 1 possible category of this paper as a python list, like ['XX']\\
    \midrule
    Ogbn-arxiv &Please predict the most appropriate arXiv Computer Science (CS) sub-category for the paper. The predicted sub-category should be in the format ['cs.XX'].\\
    \midrule
    Ogbn-products &Task: \nl There are following categories: \nl ['Home \& Kitchen', 'Health \& Personal Care', 'Beauty', 'Sports \& Outdoors', 'Books', 'Patio, Lawn \& Garden', 'Toys \& Games', 'CDs \& Vinyl', 'Cell Phones \& Accessories', 'Grocery \& Gourmet Food', 'Arts, Crafts \& Sewing', 'Clothing, Shoes \& Jewelry', 'Electronics', 'Movies \& TV', 'Software', 'Video Games', 'Automotive', 'Pet Supplies', 'Office Products', 'Industrial \& Scientific', 'Musical Instruments', 'Tools \& Home Improvement', 'Magazine Subscriptions', 'Baby Products', 'label 25', 'Appliances', 'Kitchen \& Dining', 'Collectibles \& Fine Art', 'All Beauty', 'Luxury Beauty', 'Amazon Fashion', 'Computers', 'All Electronics', 'Purchase Circles', 'MP3 Players \& Accessories', 'Gift Cards', 'Office \& School Supplies', 'Home Improvement', 'Camera \& Photo', 'GPS \& Navigation', 'Digital Music', 'Car Electronics', 'Baby', 'Kindle Store', 'Buy a Kindle', 'Furniture \& Decor', '\#508510']\nl Please predict the most likely category of this product from Amazon. Please output in the form ['your category']. \\
    \bottomrule
\end{tabular}

\end{table*}

\begin{table*}[htbp]
    \caption{The \param{Task Instruction w/ Neighbor} in Table \ref{table:prompt}.}
    \label{table:task_instruction_w}
    \begin{tabular}[]{>{\centering\arraybackslash} p{3cm} | >{\centering\arraybackslash} p{13.2cm}}
    \toprule

    & \param{Task Instruction w/ Neighbor}  \\
    \midrule
    Cora  & Task: \nl There are following categories: \nl ['Rule Learning', 'Case Based', 'Genetic Algorithms', 'Theory', 'Reinforcement Learning', 'Probabilistic Methods', 'Neural Networks']\nl Which category does this paper belong to?\nl Please comprehensively consider the information from the categories of the neighbors, and output the most 1 possible category of this paper. Please output in the form: Category: ['category']\\
    \midrule
    PubMed& Question: Does the paper involve any cases of Type 1 diabetes, Type 2 diabetes, or Experimentally induced diabetes? Please give one of either ['Type 1 diabetes'], ['Type 2 diabetes'], or ['Experimentally induced diabetes']. Please comprehensively consider the information the information from the title, abstract and neighbors, and do not give any reasoning process. Output the most 1 possible category of this paper as a python list and in the form Category: ['XX'].\\
    \midrule
    CiteSeer &Task: \nl There are following categories: \nl ['Agents', 'Machine Learning', 'Information Retrieval', 'Database', 'Human Computer Interaction', 'Artificial Intelligence']\nl Which category does this paper belong to?\nl Please comprehensively consider the information from the article and its neighbors, and output the most 1 possible category of this paper as a python list and in the form Category: ['XX']\\
    \midrule
    Ogbn-arxiv &Please comprehensively consider the information from the categories of the neighbors and predict the most appropriate arXiv Computer Science (CS) sub-category for the paper. The predicted sub-category should be in the format ['cs.XX'].\\
    \midrule
    Ogbn-products &Task: \nl There are following categories: \nl ['Home \& Kitchen', 'Health \& Personal Care', 'Beauty', 'Sports \& Outdoors', 'Books', 'Patio, Lawn \& Garden', 'Toys \& Games', 'CDs \& Vinyl', 'Cell Phones \& Accessories', 'Grocery \& Gourmet Food', 'Arts, Crafts \& Sewing', 'Clothing, Shoes \& Jewelry', 'Electronics', 'Movies \& TV', 'Software', 'Video Games', 'Automotive', 'Pet Supplies', 'Office Products', 'Industrial \& Scientific', 'Musical Instruments', 'Tools \& Home Improvement', 'Magazine Subscriptions', 'Baby Products', 'label 25', 'Appliances', 'Kitchen \& Dining', 'Collectibles \& Fine Art', 'All Beauty', 'Luxury Beauty', 'Amazon Fashion', 'Computers', 'All Electronics', 'Purchase Circles', 'MP3 Players \& Accessories', 'Gift Cards', 'Office \& School Supplies', 'Home Improvement', 'Camera \& Photo', 'GPS \& Navigation', 'Digital Music', 'Car Electronics', 'Baby', 'Kindle Store', 'Buy a Kindle', 'Furniture \& Decor', '\#508510']\nl Please predict the most likely category of this product from Amazon. Please output in the form ['your category']. \\
    \bottomrule
\end{tabular}

\end{table*}

\end{document}